\documentclass[10pt,journal,compsoc]{IEEEtran}
\usepackage{amsmath,amsfonts}
\usepackage{algorithmic}
\usepackage{algorithm}
\usepackage{array}
\usepackage[caption=false,font=normalsize,labelfont=sf,textfont=sf]{subfig}
\usepackage{textcomp}
\usepackage{stfloats}
\usepackage{url}
\usepackage{verbatim}
\usepackage{graphicx}
\usepackage{cite}
\usepackage{booktabs}

\newtheorem{theorem}{Theorem}

\begin{document}

\title{Communicability-Inspired Positional Encoding (CIPE)}

\author{Yipeng Zhang, Zhongtian Sun, Pietro Li\`o, and Kelin Xia
\thanks{This work was supported in part by the Singapore Ministry of Education Academic Research Fund Tier 1 grant RG16/23, Tier 2 grants MOE-T2EP20125-0004. (corresponding authors: Yipeng Zhang and Kelin Xia.)}
\thanks{Yipeng Zhang and Kelin Xia are with the Division of Mathematical Sciences, School of Physical and Mathematical Sciences, Nanyang Technological University, Singapore 637371, Singapore. (e-mail: yipeng001@e.ntu.edu.sg; xiakelin@ntu.edu.sg).}
\thanks{Zhongtian Sun is with the School of Computing, University of Kent, Canterbury, Kent CT2 7NZ, UK. (e-mail: z.sun-256@kent.ac.uk).}
\thanks{Pietro Li\`o is with the Department of Computer Science and Technology, University of Cambridge, 15 JJ Thomson Avenue, Cambridge CB3 0FD, UK. (e-mail: pl219@cam.ac.uk).
}
}



\maketitle

\begin{abstract}
Positional encodings (PEs) are essential for Transformers. 
Yet designing effective PEs for non-Euclidean graphs remains challenging. 
Such encodings should ideally induce an {\bf Attention-Compatible Geometry} for self-attention: not merely describing graph structure, but defining a geometry whose inner products reflect meaningful structural relatedness.
To realize this geometry, we propose Communicability-Inspired Positional Encoding (CIPE), built from communicability, a measure between pairs of nodes that aggregates contributions from paths of all lengths. 
By construction, CIPE inner products recover communicability, converting global multi-path connectivity into an attention-ready similarity geometry. 
For practical Transformer training, we introduce dimensionality alignment, mapping graph-size-dependent CIPE representations to prescribed dimensions while faithfully preserving the induced geometry.
Empirically, CIPE improves structure-agnostic Transformers by 35.5\% on average across seven benchmarks, outperforming representative PEs; it also consistently improves structure-biased graph Transformers, where competing PEs often yield only marginal benefits. 
These results position CIPE as a principled framework for attention-compatible graph positional encodings.
\end{abstract}


\section{Introduction}

\IEEEPARstart{T}{ransformers} have achieved remarkable success across domains such as natural language processing and computer vision, largely because self-attention captures long-range dependencies~\cite{vaswani2017attention,tang2018self,devlin2019bert,gillioz2020overview,dosovitskiy2020image,khan2022transformers}. 
However, self-attention is inherently permutation-invariant: without explicit positional information, it treats the input as an unordered set. 
Positional encodings (PEs) are therefore essential for injecting positional or relational structure into Transformer models, enabling attention to depend on the relative or absolute positions of input elements. 
In sequential domains, positional mechanisms have evolved from the original sinusoidal encoding~\cite{vaswani2017attention} to relation-aware designs, including relative positional representations~\cite{shaw2018self}, positional biases~\cite{raffel2020exploring,press2022train}, and rotary schemes such as Rotary Positional Encoding (RoPE)~\cite{su2024roformer}. 
These developments highlight a key principle: because self-attention operates through similarity comparisons in representation space, the induced similarity geometry is central to Transformer behavior. RoPE explicitly realizes this principle by endowing that geometry with relative positional structure, making it one of the most widely adopted positional mechanisms for sequence Transformers~\cite{dufter2022position}.

The design of positional encoding becomes substantially more challenging for non-Euclidean data in geometric deep learning, especially graphs, which lack a canonical ordering and have inherently irregular structural relations~\cite{bronstein2021geometric}. Early graph encodings used simple node-level descriptors, such as degrees or related local statistics, to provide basic structural information~\cite{shervashidze2011weisfeiler,henderson2011s,henderson2012rolx}. Subsequent methods introduced richer graph structural encodings, including random-walk-based and cycle-based schemes~\cite{dwivedi2023benchmarking,canturk2023graph,yan2024cycle}. Another major line of work developed spectral and diffusion-based positional or structural encodings, represented by Laplacian eigenvector-based encodings such as LapPE~\cite{dwivedi2020generalization,huang2023stability} and heat-kernel-based encodings such as HKSE~\cite{kreuzer2021rethinking,mialon2021graphit}. More recent graph Transformer frameworks further combine or learn structural and positional information, as in GPS~\cite{rampavsek2022recipe} and GPSE~\cite{canturk2023graph}. Together with other studies on graph positional and structural encodings~\cite{min2022transformer,dufter2022position,corso2024graph}, these developments reveal a rich landscape of methods for encoding graph structure. 
Nevertheless, most existing methods still capture only partial aspects of graph structure. 
Few encodings are designed to reflect the full multi-path organization of a graph in a unified manner. 
Moreover, even when rich structural information is encoded, it is rarely organized into a representation geometry in which inner-product similarity directly reflects meaningful graph structural relatedness and is therefore directly usable by self-attention.

\begin{figure*}[t]
    \centering
    \includegraphics[width=\linewidth]{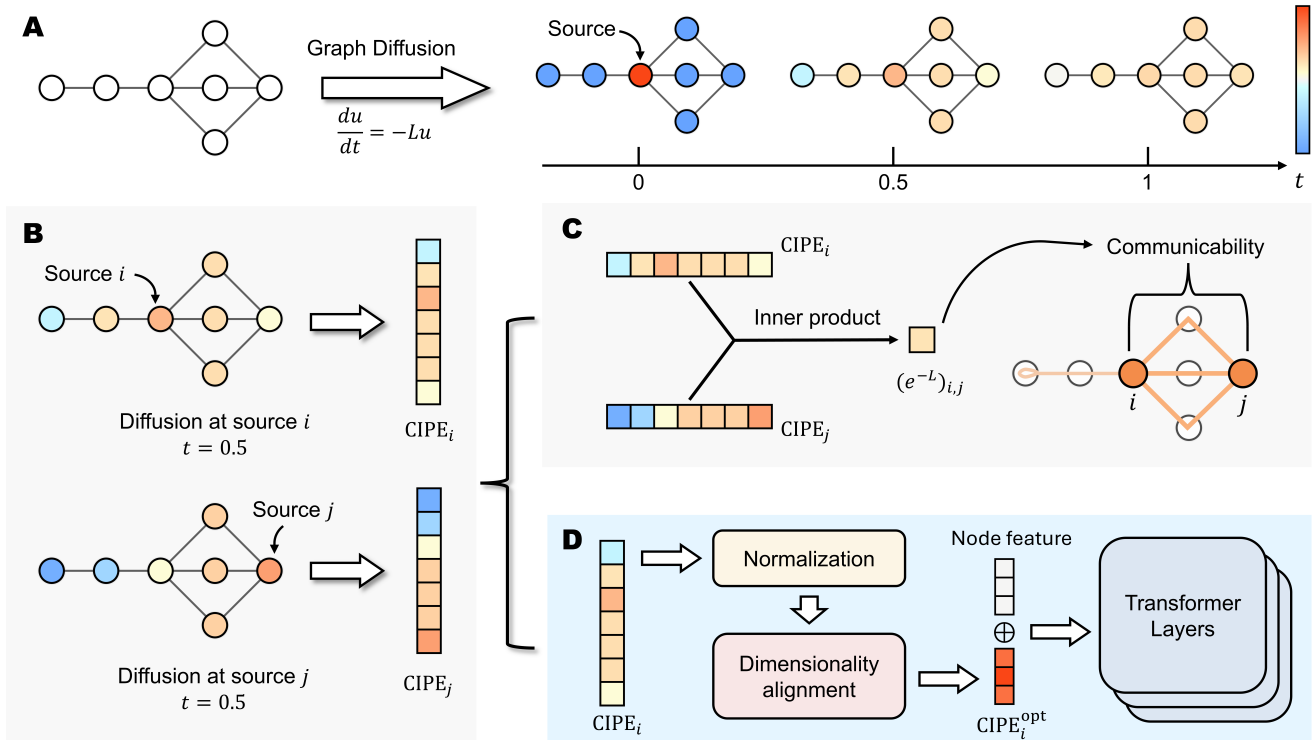}
\caption{\textbf{CIPE turns diffusion-based communicability into an attention-compatible positional geometry for graph Transformers.}
    \textbf{A} Heat diffusion on a graph initialized from a unit source, illustrating how node-wise signal propagates over the graph.
    \textbf{B} For each node, CIPE is constructed from the graph-wide diffusion profile obtained by placing a unit heat source at that node and evolving the diffusion process for time $t/2$.
    \textbf{C} The key property of CIPE: the inner product between two node encodings directly recovers their diffusion-based communicability, a pairwise quantity that aggregates contributions from all possible connecting paths between the two nodes with different weights according to path length. This satisfies the \textbf{Attention-Compatible Geometry} condition by making the similarity computation used by self-attention directly reflect graph structural relatedness.
    \textbf{D} For practical deployment, CIPE is normalized and dimensionally aligned into a prescribed shared embedding space while preserving its communicability-induced inner-product geometry as faithfully as possible, then concatenated with node features and fed into Transformer layers for graph representation learning.}
    \label{fig:CIPE_flowchart}
\end{figure*}

A key design question in graph positional encoding is which graph relation should serve as the basis for positional representation. In this regard, communicability is particularly appealing. 
It has long been recognized as a fundamental graph descriptor, measuring how easily information propagates through a network, and has proved effective in graph-analytic tasks such as clustering, vertex similarity, and role discovery~\cite{leicht2006vertex,pereda2019visualization,estrada2024communicability}. 
Intuitively, communicability aggregates contributions from all possible connecting paths between two vertices, rather than relying only on adjacency or shortest-path proximity. 
Originally introduced in network science~\cite{estrada2008communicability,estrada2011structure}, it combines local and global connectivity through length-weighted path contributions. 
This enables it to capture the broader potential for information transmission and structural influence across the graph. 
Communicability-based embeddings further reveal meaningful geometric and functional organization in complex systems, linking communicability to network efficiency, robustness, and higher-order coordination~\cite{grindrod2011communicability,estrada2014hyperspherical, estrada2012physics,estrada2016communicability,pereda2019visualization}. 
These observations suggest that communicability provides a natural foundation for graph positional encodings that capture globally distributed, multi-path structure.

Motivated by these observations, we propose {Communicability-Inspired Positional Encoding} ({CIPE}) for graph Transformers. 
Rather than introducing another structure descriptor, CIPE constructs a positional geometry over graph nodes whose inner products reflect pairwise communicability, making the induced similarity geometry naturally compatible with self-attention. 
We refer to this design principle as the {\bf Attention-Compatible Geometry} condition. 
Starting from the discrete diffusion equation on graphs, CIPE is derived from the analytical solution of the associated graph operator, yielding node encodings that combine local connectivity and long-range multi-path interactions in a common similarity geometry. 
Its key property is that the inner product between the positional encodings of two vertices recovers their diffusion-based communicability, so node pairs with stronger multi-path structural relatedness have higher positional similarity. 
Thus, CIPE organizes global graph structure in the same inner-product geometry used by attention. 
In particular, when specialized to one-dimensional sequences, the CIPE construction exhibits a desirable property on periodic sequences, equivalently cycle graphs: its inner products, similar to RoPE, depend only on relative positional offsets. 
This suggests potential future relevance to intrinsically periodic sequence representations, such as polymer SMILES.

A central practical challenge is that raw CIPE is graph-size dependent, whereas Transformer architectures require positional representations from different graphs to lie in a common embedding space. 
To bridge this mismatch, we introduce a dimensionality alignment procedure based on inner-product preservation, which maps CIPE into a prescribed dimension while minimizing distortion of its communicability-induced similarity geometry. 
Thus, CIPE can be compressed or expanded for downstream requirements while preserving the geometry most relevant to self-attention as faithfully as possible. 
Figure~\ref{fig:CIPE_flowchart} illustrates the communicability-driven construction of CIPE and its integration into Transformer-based graph learning.

Extensive experiments on {14} benchmarks---including {ten} molecular property prediction tasks from MoleculeNet~\cite{wu2018moleculenet} (seven classification and three quantum-chemistry regression datasets) and {four} non-molecular graph classification benchmarks from TUDataset~\cite{morris2020tudataset} (covering social-network and bio/protein graphs)---demonstrate that CIPE consistently outperforms existing positional encodings. In particular, CIPE delivers especially pronounced improvements on structure-agnostic Transformer backbones, while continuing to provide additional gains even when incorporated into graph Transformer architectures that already encode structural bias. Moreover, despite using only a few Transformer layers, the resulting models remain competitive with widely used graph learning baselines. Taken together, these findings suggest that communicability offers not only a powerful description of graph structure, but also a principled foundation for designing positional geometry naturally compatible with attention. 
The main contributions of this work are summarized as follows:
\begin{itemize}
    \item We introduce CIPE, a graph positional encoding that leverages communicability to construct a globally informative positional geometry naturally compatible with the self-attention mechanism in Transformers.
    \item We develop a stable dimensionality alignment framework that maps CIPE to a prescribed embedding dimension while preserving pairwise inner products as faithfully as possible, thereby enabling its practical use on graphs of varying sizes.
    \item We demonstrate across diverse graph types, learning tasks, and Transformer backbones that CIPE consistently improves performance, with gains that substantially exceed those of existing positional encodings.
\end{itemize}

\section{Related Work}

\subsection{Positional Encodings for Sequence Data}

For sequence data, positional encoding has been extensively studied as a mechanism for incorporating order information into self-attention. 
The original Transformer introduced fixed sinusoidal positional encodings, which inject absolute token positions by adding deterministic trigonometric vectors to input embeddings~\cite{vaswani2017attention}. 
This design gives each position a distinct representation and allows the model to use positional information without recurrent or convolutional structures. 
However, because the positional signal is supplied only through the input embeddings, the attention operation itself is not explicitly parameterized by the relative distance between two tokens.

Subsequent works therefore moved from absolute position injection toward relation-aware attention mechanisms. 
Shaw~\emph{et~al.}~\cite{shaw2018self} extended self-attention with relative position representations, enabling attention to explicitly consider the distance between sequence elements rather than relying only on absolute positional vectors. 
This changes positional encoding from a token-wise input feature into a pairwise relation used inside the attention computation. 
Bias-based methods further simplify this idea by adding position-dependent terms directly to attention logits. 
For example, T5 uses relative position biases in its text-to-text Transformer framework~\cite{raffel2020exploring}, while ALiBi adds linear distance-dependent biases to query-key attention scores, allowing models trained on shorter sequences to extrapolate to longer ones~\cite{press2022train}. 
These methods show that positional information can be injected not only through node or token representations, but also by directly modifying the similarity scores used by attention.

Rotary Positional Encoding (RoPE) provides a particularly relevant example for this work. 
Instead of adding a positional vector to token embeddings or adding a scalar bias to attention logits, RoPE encodes positions through rotations applied to query and key representations~\cite{su2024roformer}. 
As a result, the inner product between rotated query and key vectors becomes explicitly dependent on their relative positional offset. 
This makes RoPE more than a positional descriptor: it induces an inner-product geometry in which the similarity computation of self-attention reflects relative sequence structure. 
This perspective is also consistent with broader analyses showing that positional information in Transformers can be understood through how different encoding schemes modify the geometry and computation of attention~\cite{dufter2022position}. 
In this sense, RoPE represents an important example of an attention-compatible positional geometry for ordered sequence data.

\subsection{Positional Encodings for Graph Data}

Early positional encodings for graph data used simple node-level descriptors, such as degrees, local neighborhood statistics, or recursively aggregated structural features~\cite{shervashidze2011weisfeiler,henderson2011s,henderson2012rolx}. 
The main idea of these designs is to attach basic structural attributes to each node, so that the model can distinguish nodes not only by their original features but also by their local structural roles.

To incorporate richer graph structural information into node or graph encodings, later works introduced structural encodings (SEs), which characterize graphs through graph-derived structural patterns. 
Cycle-counting encodings, such as CycleSE, use the number of cycles of different lengths as structural signals, aiming to capture higher-order topological patterns beyond local neighborhoods~\cite{canturk2023graph,yan2024cycle}. 
Random Walk Structural Encoding (RWSE) instead describes a node through multi-step random-walk behavior, especially the probabilities that walks starting from the node return to it at different lengths~\cite{dwivedi2023benchmarking}. 
In this way, RWSE provides a diffusion-like structural signature based on random-walk dynamics. 
Heat-kernel-based structural encodings, such as HKSE, follow a related diffusion perspective but use heat propagation on graphs to summarize how information spreads from each node across multiple scales~\cite{kreuzer2021rethinking,mialon2021graphit}.

Another classical direction is based on spectral graph representations. 
Laplacian eigenvector-based positional encodings, such as LapPE, construct node coordinates from the eigenvectors of the graph Laplacian~\cite{dwivedi2020generalization}. 
This makes LapPE conceptually analogous to sinusoidal positional encodings for sequences: both use frequency-like components to provide positional information, but LapPE replaces fixed sequence frequencies with graph-dependent spectral components. 
At the same time, spectral graph encodings are known to suffer from sign ambiguity, basis non-uniqueness, and sensitivity to structural perturbations~\cite{huang2023stability}.

More recent works have moved from manually specified graph encodings toward frameworks that integrate or learn positional and structural information. 
GPS treats positional and structural encodings as an explicit module in a scalable graph Transformer architecture, showing that the choice of PE/SE is an important component when combining local message passing with global attention~\cite{rampavsek2022recipe}. 
GPSE is more directly focused on encoding design: it learns transferable positional and structural representations through pretraining objectives, rather than relying on a single hand-crafted encoding for each task~\cite{canturk2023graph}. 
Together with other studies on graph positional and structural encodings~\cite{min2022transformer,dufter2022position,corso2024graph}, these developments reveal a rich landscape of methods for injecting graph structure into Transformer-based graph learning.

Overall, existing graph PEs and SEs have progressively enriched node representations with local, global, spectral, diffusion-based, and learned structural information. 
This line of work has greatly expanded the ways in which graph structure can be encoded for downstream learning. 
However, most existing designs are descriptor-oriented: they focus on what structural information should be encoded, but not on how the resulting positional geometry interacts with the inner-product similarity used by self-attention. 
From the perspective of graph Transformers, this leaves a mismatch between graph structural encoding and the geometry of the attention mechanism. 
This motivates the attention-compatible geometry principle developed in this work, where positional encodings are constructed so that their inner products directly realize a meaningful graph-theoretic measure of structural relatedness.

\subsection{Communicability on Graphs}

Communicability is a classical graph-theoretic concept for measuring how effectively two vertices are connected through the network~\cite{estrada2008communicability,estrada2011structure}. 
Its key idea is to move beyond adjacency and shortest-path distance by accounting for the collective contribution of walks connecting two vertices. 
Short walks capture strong local proximity, while longer walks still contribute to indirect structural influence with smaller weights. 
As a result, communicability provides a unified measure that combines local connectivity, indirect interactions, and global multi-path organization.

This all-path perspective has made communicability useful for analyzing complex networks. 
It has been used to characterize vertex similarity, clustering, role discovery, and the potential for information transmission across networks~\cite{leicht2006vertex,pereda2019visualization,estrada2024communicability}. 
Communicability graph methods further use pairwise communicability to reveal community structures, where groups of vertices are identified by stronger internal communicability than external communicability~\cite{estrada2009communicabilitygraph}. 
Related extensions, such as communicability betweenness, quantify how much a node participates in communication between other pairs of nodes by considering information flow through all possible routes rather than only shortest paths~\cite{estrada2009communicability}.

Communicability has also led to geometric interpretations of graph structure. 
Communicability distance defines a distance-like measure between vertices by comparing their self-communicability and mutual communicability, enabling networks to be studied in Euclidean communicability spaces~\cite{estrada2012complex}. 
Hyperspherical embeddings of graphs and networks in communicability spaces further show that communicability can induce meaningful geometric representations of network organization~\cite{estrada2014hyperspherical}. 
Communicability angle provides another geometric view, relating pairwise communicability to the spatial efficiency of networks and revealing structural organization in a variety of real-world systems~\cite{estrada2016communicability}. 
These studies suggest that communicability is not merely a scalar graph statistic, but a rich structural principle connecting paths, information flow, similarity, and geometry.

Communicability provides a principled measure of pairwise structural relatedness, reflecting both local connectivity and globally distributed multi-path interactions. 
Motivated by this perspective, we design CIPE to translate communicability-based structural relatedness into node-wise positional encodings. 
Specifically, CIPE constructs a positional geometry in which the inner product between two node encodings directly reflects the communicability-quantified relation between the corresponding vertices.

\section{METHODOLOGY}
\subsection{Preliminary}
\subsubsection{Graph Laplacian}

Let $G=(V,E)$ be an undirected graph with $n$ vertices, adjacency matrix $A\in\{0,1\}^{n\times n}$, and degree matrix $D=\mathrm{diag}(d_1,\dots,d_n)$, where $d_i=\sum_{j=1}^n A_{ij}$. The combinatorial graph Laplacian is defined as
\begin{equation}
L = D - A.
\end{equation}
This operator serves as the discrete analogue of the Laplace operator and underlies diffusion dynamics on the graph.

\subsubsection{Diffusion Equation on Graphs}

Given an initial signal $u_0\in\mathbb{R}^n$ on the vertices of $G$, the continuous-time diffusion process on the graph is governed by
\begin{equation}
\frac{d}{dt}u(t) = -L\,u(t), \qquad u(0)=u_0.
\end{equation}
Its analytical solution is
\begin{equation}
u(t)=e^{-tL}u_0,
\end{equation}
where $e^{-tL}$ is the {graph heat kernel}.

In particular, if the initial signal is a unit impulse at vertex $v_j$, i.e., $u(0)=\delta_j$, then
\[
u(t)=e^{-tL}\delta_j,
\]
and the amount of signal at vertex $v_i$ at time $t$ is
\[
u_i(t)=\delta_i^\top e^{-tL}\delta_j=(e^{-tL})_{ij}.
\]
Thus, the entry $(e^{-tL})_{ij}$ quantifies how much signal reaches vertex $v_i$ from an initial unit mass placed at vertex $v_j$ after diffusion time $t$.

\subsubsection{Communicability from a Diffusion Perspective}
\label{sec:communicability}

The heat kernel $e^{-tL}$ provides a natural way to quantify how strongly two vertices are related under graph diffusion. In particular, the entry $(e^{-tL})_{ij}$ measures how much signal reaches vertex $v_i$ at time $t$ when the process starts from a unit impulse at vertex $v_j$. Unlike shortest-path distance or purely local structural descriptors, this quantity reflects the cumulative effect of propagation through the graph as a whole.

Motivated by this interpretation, we define the diffusion-based communicability between vertices $v_i$ and $v_j$ as
\begin{equation}
F_{ij}:=(e^{-tL})_{ij}.
\end{equation}
This quantity serves as a global measure of how efficiently information can propagate between the two vertices under the diffusion process.

To remove the effect of absolute scale and focus on relative relatedness, we further introduce the communicability cosine similarity~\cite{estrada2024communicability}:
\begin{equation}
\cos\theta_{ij}:=\frac{F_{ij}}{\sqrt{F_{ii}F_{jj}}}.
\end{equation}
This normalized quantity measures the angular similarity between vertices in the diffusion-induced space. A value of $\cos\theta_{ij}$ close to $1$ indicates that $v_i$ and $v_j$ are strongly related relative to their own self-communicability, and therefore occupy closely aligned positions in the global diffusion geometry of the graph.

\subsection{Communicability-Inspired Positional Encoding (CIPE)}

We now construct a positional encoding whose inner-product geometry directly realizes diffusion-based communicability. For a graph $G$ with Laplacian $L$, we define the {Communicability-Inspired Positional Encoding} (CIPE) of vertex $v_i$ as
\begin{equation}
\mathrm{CIPE}_i = e^{-\tfrac{t}{2}L}\delta_i \in \mathbb{R}^n,
\end{equation}
where $\delta_i$ is the Kronecker delta vector supported on vertex $v_i$. Equivalently, $\mathrm{CIPE}_i$ is the diffusion state at time $t/2$ generated by a unit impulse placed at vertex $v_i$, and each coordinate records how much signal has reached the corresponding vertex after half the diffusion time.

The choice of $t/2$ is essential rather than arbitrary. Indeed, for any two vertices $v_i$ and $v_j$, we have
\begin{equation}
\langle \mathrm{CIPE}_i, \mathrm{CIPE}_j \rangle
= \delta_i^\top e^{-\tfrac{t}{2}L} e^{-\tfrac{t}{2}L} \delta_j
= \delta_i^\top e^{-tL}\delta_j
= (e^{-tL})_{ij}.
\end{equation}
Thus, the inner product between $\mathrm{CIPE}_i$ and $\mathrm{CIPE}_j$ exactly recovers the diffusion-based communicability between $v_i$ and $v_j$ at time $t$. In this way, CIPE turns communicability into a positional geometry over nodes, making structural relatedness directly usable through inner-product similarity.

\subsection{Desirable Properties of CIPE}

\begin{theorem}\label{thm:properties}
For a graph $G$ with Laplacian $L$ and communicability-inspired positional encoding defined as above, the following properties hold:
\begin{enumerate}
    \item \textbf{Discriminativity:}
    \[
    \mathrm{CIPE}_i \neq \mathrm{CIPE}_j \quad \text{for } i\neq j.
    \]

    \item \textbf{Attention-compatible geometry:}
    \[
    \langle \mathrm{CIPE}_i, \mathrm{CIPE}_j \rangle = F(v_i,v_j),
    \]
    where $F(v_i,v_j)$ is a function that captures a meaningful structural relation between vertices $v_i$ and $v_j$.

    \item \textbf{Stability:} For any two graphs $G$ and $G'$ that differ by a small perturbation of their structure, there exists $M>0$ such that
    \[
    \|\mathrm{CIPE}_i(G') - \mathrm{CIPE}_i(G)\| \le M \cdot d(G',G).
    \]
    Here, $d(G',G)$ denotes a metric that quantifies the structural discrepancy between the two graphs.
\end{enumerate}
\end{theorem}

These three properties highlight why CIPE is well suited as a positional encoding mechanism for graph-structured data. \textbf{Discriminativity} ensures that different vertices are assigned distinct encodings, so that the Transformer can distinguish nodes based on position rather than collapsing them into indistinguishable embeddings. \textbf{Attention-compatible geometry} means that the inner product between two positional encodings reflects a meaningful structural relation between the corresponding vertices, making the encoding naturally compatible with the similarity computation underlying self-attention. In the case of CIPE, this structural relation is instantiated by diffusion-based communicability, namely
\[
F(v_i,v_j) = (e^{-tL})_{ij}.
\]
Finally, \textbf{Stability} guarantees that small perturbations of the graph structure induce only bounded changes in the positional encoding, which is important for robustness in practical applications.

\subsection{Normalization of CIPE}

The raw CIPE matrix records absolute diffusion responses and is therefore influenced by graph size and density. To obtain a scale-independent representation of relative communicability, we normalize CIPE at the matrix level using a symmetric normalization scheme. Let
\[
\mathrm{CIPE}=e^{-\tfrac{t}{2}L}\in\mathbb{R}^{n\times n},
\]
where the $i$-th row corresponds to the encoding $\mathrm{CIPE}_i$ of vertex $v_i$. We define the normalized encoding matrix as
\begin{equation}
\widehat{\mathrm{CIPE}}
= D^{-\tfrac{1}{2}}\,\mathrm{CIPE}\,D^{-\tfrac{1}{2}},
\qquad
D_{ii}=(\mathrm{CIPE})_{ii}=(e^{-\tfrac{t}{2}L})_{ii}.
\end{equation}

Under this normalization, each entry of $\widehat{\mathrm{CIPE}}$ becomes
\[
(\widehat{\mathrm{CIPE}})_{ij}
=
\frac{(e^{-\tfrac{t}{2}L})_{ij}}
{\sqrt{(e^{-\tfrac{t}{2}L})_{ii}(e^{-\tfrac{t}{2}L})_{jj}}}
=
\cos\theta_{ij},
\]
which is exactly the communicability cosine introduced in Section~\ref{sec:communicability}.

Thus, normalization converts CIPE from an absolute diffusion-response encoding into a relative communicability geometry: instead of measuring raw signal magnitude, each entry now reflects the angular relation between two vertices in the diffusion-induced space. In this form, $\widehat{\mathrm{CIPE}}$ directly encodes the relative communicability structure of the graph, enabling consistent comparison across graphs of different sizes and densities.

\subsection{Dimensionality Alignment of CIPE}
\label{sec:reshape}

A fundamental practical challenge is that raw CIPE has graph-dependent dimensionality equal to the number of vertices. This is incompatible with parameter-sharing architectures such as Transformers, which require positional encodings from different graphs to lie in a common embedding space. We therefore introduce a dimensionality alignment procedure that maps CIPE into a prescribed dimension $d$ while preserving its communicability-induced inner-product geometry as faithfully as possible.

Let the normalized positional encoding matrix be $\widehat{\mathrm{CIPE}}$, whose $i$-th row is the normalized encoding of vertex $v_i$. We seek a fixed-dimensional encoding matrix
\[
\mathrm{CIPE}^{(d)} \in \mathbb{R}^{n\times d},
\]
such that pairwise inner products between rows of \(\mathrm{CIPE}^{(d)}\) approximate those of \(\widehat{\mathrm{CIPE}}\). Equivalently, we solve
\begin{equation}
\min_{\mathrm{CIPE}^{(d)}\in\mathbb{R}^{n\times d}}
\; \big\| \mathrm{CIPE}^{(d)}{\mathrm{CIPE}^{(d)}}^\top
- \widehat{\mathrm{CIPE}}\,\widehat{\mathrm{CIPE}}^\top \big\|_F,
\label{eq:align_obj}
\end{equation}
so that
\[
\langle \mathrm{CIPE}^{(d)}_i,\mathrm{CIPE}^{(d)}_j\rangle
\approx
\langle \widehat{\mathrm{CIPE}}_i,\widehat{\mathrm{CIPE}}_j\rangle.
\]
This formulation makes explicit that the goal is not merely dimensionality reduction, but the faithful transfer of communicability-induced geometry into a fixed-dimensional representation.

Let the eigendecomposition of the target Gram matrix be
\[
\widehat{\mathrm{CIPE}}\,\widehat{\mathrm{CIPE}}^\top = U \Sigma U^\top,
\]
where \(U=(u_1,\dots,u_n)\in\mathbb{R}^{n\times n}\) is orthogonal and
\(\Sigma=\mathrm{diag}(\sigma_1,\dots,\sigma_n)\) with \(\sigma_1\ge\cdots\ge\sigma_n\ge0\).
The rank-\(d\) family of minimizers of (\ref{eq:align_obj}) is
\[
{\mathrm{CIPE}^{(d)}}^* = U_d \Sigma_d^{1/2} D,
\]
where \(U_d\in\mathbb{R}^{n\times d}\) contains the top-\(d\) eigenvectors,
\(\Sigma_d\in\mathbb{R}^{d\times d}\) contains the corresponding eigenvalues,
and \(D\in\mathbb{R}^{d\times d}\) is any orthogonal matrix. If \(d>n\), we zero-pad \(U_d\) and \(\Sigma_d\) to match the target dimension.

To promote a more uniform distribution of information across the output dimensions, we first introduce a near-orthogonal projection
\begin{equation}
    P_{n,d} = O_n J_{n,d} O_d \in\mathbb{R}^{n\times d},
    \label{eq:near_orth_mat}
\end{equation}

where \(O_n\) and \(O_d\) are orthogonal DCT matrices of sizes \(n\times n\) and \(d\times d\), respectively, and \(J_{n,d}\) is the \(n\times d\) diagonal matrix with ones on the main diagonal. We then choose the orthogonal matrix \(D^{\mathrm{opt}}\) that best aligns \(U_d\Sigma_d^{1/2}\) with the projected data \(\widehat{\mathrm{CIPE}}P_{n,d}\) by solving
\[
\min_{D\in\mathcal{O}(d)} \big\| U_d \Sigma_d^{1/2} D - \widehat{\mathrm{CIPE}} P_{n,d} \big\|_F.
\]
If
\[
\big(U_d\Sigma_d^{1/2}\big)\big(\widehat{\mathrm{CIPE}} P_{n,d}\big)^\top = X \Gamma Y^\top
\]
is the corresponding singular value decomposition, then the optimal orthogonal matrix is \(D^{\mathrm{opt}}=XY^\top\), yielding the refined encoding
\begin{equation}
\mathrm{CIPE}^{\mathrm{opt}} = U_d \Sigma_d^{1/2} X Y^\top \in\mathbb{R}^{n\times d}.
\end{equation}

The resulting \(\mathrm{CIPE}^{\mathrm{opt}}\) serves as the fixed-dimensional positional encoding supplied to downstream Transformer layers. In this way, dimensionality alignment makes CIPE deployable across graphs of varying sizes while preserving the original inner-product geometry as closely as possible in the prescribed dimension.

\subsection{Properties Preserved After Dimensionality Alignment}

The dimensionality alignment procedure is designed to preserve the favorable properties of CIPE as much as possible when projecting it into a fixed-dimensional space. A complete theorem together with its proof is provided in the Supplementary Information. Here, we summarize the main implications.

First, when the target dimension satisfies $d\ge n$, the aligned encoding retains all properties of the original CIPE exactly. In particular, discriminativity, attention-compatible geometry, and stability are preserved without loss. The nontrivial case is therefore the practically relevant regime $d<n$, where compression is unavoidable.

In this compressed regime, the most important property preserved by construction is the inner-product geometry. Specifically, the aligned encoding $\mathrm{CIPE}^{\mathrm{opt}}$ preserves the pairwise inner products of the normalized CIPE as well as any $n\times d$ embedding possibly can, in the sense of minimizing the Frobenius error between Gram matrices. Thus, although exact preservation is generally impossible when $d<n$, the communicability-induced geometry is retained in an optimal sense within the prescribed dimension.

Beyond this optimal inner-product preservation, discriminativity is retained almost everywhere for $d>1$, so that distinct vertices remain distinguishable in generic cases after alignment. Moreover, under a standard spectral gap condition $\sigma_d>\sigma_{d+1}$, the aligned encoding is independent of arbitrary eigenvector choices within the top-$d$ subspace and varies continuously under small perturbations of the graph. This yields a stable fixed-dimensional positional encoding that avoids the eigenvector ambiguity issues often associated with classical spectral positional encodings.

Taken together, these observations show that dimensionality alignment does not merely compress CIPE, but transfers its communicability-induced positional geometry into a shared embedding space while retaining the properties most relevant for downstream Transformer deployment.

\subsection{Scalable Approximation for Large Graphs}

For a graph $G$ with $n$ vertices, the exact construction of CIPE becomes expensive on large graphs. Similar to other spectral positional encodings, both the computation of the heat kernel $e^{-tL}$ and the dimensionality alignment step involve eigendecompositions of $n\times n$ matrices, leading in general to an overall time complexity of order $O(n^3)$. This substantially limits the scalability of the exact formulation.

To extend CIPE to larger graphs, we introduce a deterministic low-dimensional approximation based on the near-orthogonal matrix $P_{n,d}$ defined in ~\eqref{eq:near_orth_mat}. Instead of constructing the full graph-size-dependent CIPE, we use
\[
\widetilde{\mathrm{CIPE}}^{(d)} := e^{-tL/2}P_{n,d}\in\mathbb{R}^{n\times d}
\]
as the large-graph positional representation. Since $P_{n,d}$ approximately preserves orthogonality, the resulting inner-product matrix still preserves the heat-kernel-induced similarity geometry to a considerable extent, while directly producing a fixed $d$-dimensional positional encoding.

To compute $e^{-tL/2}P_{n,d}$ efficiently, we do not explicitly form the full matrix $e^{-tL/2}$ and then multiply it by $P_{n,d}$. Instead, we approximate this block heat-kernel action directly using a Chebyshev polynomial approximation of the heat operator, following the fast spectral filtering strategy in \cite{hammond2011wavelets,al2011computing}. In this way, the computation reduces to a sequence of sparse matrix--thin matrix multiplications involving $L$ and matrices of size $n\times d$.

Under this approximation, constructing $P_{n,d}$ costs $O(nd)$, while evaluating the degree-$K$ Chebyshev approximation requires $O(Kmd)$ operations for a graph with $m$ edges. Therefore, the overall time complexity becomes
\[
O(nd + Kmd),
\]
which reduces to $O(Knd)$ for sparse graphs with $m=O(n)$. When $d$ is fixed and $K$ remains moderate, this is substantially lower than the $O(n^3)$ complexity of the exact spectral construction, making the approximation suitable for large-scale graph positional encoding.

\section{EXPERIMENTS}

\subsection{Experimental protocol and baselines}\label{sec:protocol_baselines}

To separate the structural contribution of positional encoding from its interaction with existing graph inductive bias, we design two complementary experimental settings. 
First, we add positional encodings to a {structure-agnostic} Vanilla Transformer ({VTr}) backbone, which contains neither message passing nor graph-specific modules, so that performance gains mainly reflect the structural signal introduced by the encoding. 
Second, we add positional encodings to a {structure-biased} Message-Passing Transformer ({MPTr}) backbone, where a single GINE layer provides local message-passing bias before the Transformer layers, to test whether CIPE remains beneficial when structural inductive bias is already present.

We evaluate on 14 benchmarks spanning graph classification and molecular property prediction: four TUDataset graph classification datasets (IMDB-B, IMDB-M, PROTEINS, ENZYMES), three molecular regression datasets (QM7, QM8, QM9), and seven MoleculeNet molecular classification datasets (BACE, BBBP, ClinTox, SIDER, Tox21, HIV, MUV). 
We report accuracy for TUDataset, MAE for QM7/8/9, and ROC-AUC for MoleculeNet, averaged across tasks for multi-task datasets.

We compare CIPE with representative positional and structural encoding baselines, including LapPE, RWSE, CycleSE, and HKSE. 
All encodings are projected to the same dimensionality and injected through the same scheme, namely concatenation to node features before the Transformer, ensuring a controlled comparison.

Beyond these controlled PE comparisons, we include representative MoleculeNet baselines to contextualize overall model competitiveness under the same scaffold-split protocol. These baselines cover two mainstream families of graph learning models: {GNN-based} methods that rely on message passing, and {graph Transformer-based} methods that model global interactions via self-attention on graphs. The GNN-based baselines include widely used self-supervised pre-training paradigms, including SSL pre-training~\cite{hu2020strategies}, GraphCL~\cite{you2020graph}, InfoGraph~\cite{wang2023evaluating}, JOAOv2~\cite{wang2023evaluating}, GraphMAE~\cite{hou2022graphmae}, and GraphLoG~\cite{xu2021self}. The graph Transformer-based baselines include GraphGPS~\cite{rampavsek2022recipe}, AttentiveFP~\cite{xiong2019pushing}, and G2PT~\cite{chen2025graph}. In addition, we consider GPSE$_{\text{augmented}}$~\cite{canturk2024graph}, which incorporates a self-supervised learned structural encoding as an auxiliary signal for downstream prediction.

Further implementation details, including full feature specifications and hyperparameter configurations, are provided in the Supplementary Information, Section 1 and Tables S1 and S2.

\subsection{Positional encoding on Structure-Agnostic Transformer}

\begin{table*}[ht] 
\centering
\caption{Performance of positional encoding on the structure-agnostic vanilla Transformer (VTr). Results are reported as mean $\pm$ standard deviation over 5 runs. For TUDataset we report accuracy (ACC, $\uparrow$) and for QM7/8/9 we report mean absolute error (MAE, $\downarrow$). The last row shows relative improvement of VTr+CIPE over VTr. Best results are highlighted in bold.}
\label{tab:vtr_pe}
\resizebox{\linewidth}{!}{
\begin{tabular}{@{}lccccccc@{}}
\toprule
Model & IMDB-B & IMDB-M & PROTEINS & ENZYMES & QM7 & QM8 & QM9 \\
Metric & \multicolumn{4}{c}{ACC($\uparrow$)} & \multicolumn{3}{c}{MAE($\downarrow$)} \\
\cmidrule(lr){2-5}\cmidrule(lr){6-8}
No. Graphs & 1000 & 1500 & 1133 & 600 & 6830 & 21786 & 133885 \\
Avg. nodes & 19.77 & 13.00 & 39.06 & 32.63 & 16 & 16 & 18 \\
No. tasks & 1 & 1 & 1 & 1 & 1 & 12 & 16 \\
\midrule
VTr & 0.500$_{(0.000)}$ & 0.333$_{(0.000)}$ & 0.749$_{(0.020)}$ & 0.347$_{(0.014)}$ &
72.4$_{(1.0)}$ & 0.0222$_{(0.0002)}$ & 0.01207$_{(0.00009)}$ \\
VTr+LapPE & 0.512$_{(0.024)}$ & 0.488$_{(0.020)}$ & 0.787$_{(0.012)}$ & 0.313$_{(0.052)}$ &
70.4$_{(1.7)}$ & 0.0223$_{(0.0003)}$ & 0.01032$_{(0.00007)}$ \\
VTr+RWSE & 0.584$_{(0.105)}$ & 0.514$_{(0.010)}$ & 0.802$_{(0.008)}$ & 0.437$_{(0.032)}$ &
70.2$_{(1.9)}$ & 0.0215$_{(0.0003)}$ & 0.00889$_{(0.00011)}$ \\
VTr+CycleSE & 0.608$_{(0.008)}$ & 0.529$_{(0.003)}$ & 0.788$_{(0.016)}$ & 0.327$_{(0.025)}$ &
68.2$_{(0.4)}$ & 0.0217$_{(0.0002)}$ & 0.01036$_{(0.00010)}$ \\
VTr+HKSE & 0.561$_{(0.067)}$ & 0.498$_{(0.006)}$ & 0.793$_{(0.022)}$ & 0.453$_{(0.021)}$ &
67.9$_{(1.5)}$ & 0.0210$_{(0.0003)}$ & 0.00888$_{(0.00012)}$ \\
\midrule
VTr+CIPE (ours) & \textbf{0.742}$_{(0.028)}$ & \textbf{0.541}$_{(0.005)}$ & \textbf{0.822}$_{(0.017)}$ & \textbf{0.609}$_{(0.035)}$ &
\textbf{63.8}$_{(1.1)}$ & \textbf{0.0206}$_{(0.0003)}$ & \textbf{0.00805}$_{(0.00012)}$ \\
Improvement (\%) & +48.4\% & +62.5\% & +9.7\% & +75.5\% & +11.9\% & +7.2\% & +33.3\% \\
\bottomrule
\end{tabular}
}
\end{table*}

We first evaluate positional encodings on the structure-agnostic Vanilla Transformer (VTr), where performance differences primarily reflect the structural signal injected by the encoding itself. As shown in Table~\ref{tab:vtr_pe}, CIPE delivers the strongest performance on all seven benchmarks, consistently outperforming both the vanilla model and all competing positional/structural encodings. This setting is particularly informative because VTr contains no message passing, edge-aware operations, or other graph-specific inductive biases; consequently, performance differences primarily reflect how effectively each encoding injects structural information into self-attention.

Overall, CIPE delivers not only the strongest but also the most consistently superior gains among all tested positional/structural encodings. On the four TUDataset graph classification benchmarks, CIPE improves accuracy over VTr by an average of 49.0\%, with especially pronounced gains on IMDB-B, IMDB-M, and ENZYMES. On the QM7/8/9 molecular regression tasks, CIPE reduces MAE by an average of 17.5\%, with the largest reduction observed on QM9. Under this controlled setting, performance gains depend largely on the structural signal contributed by the positional encoding and on how naturally that signal interfaces with the Transformer backbone. The superior performance of CIPE therefore suggests that it is the most effective among the tested encodings in both injecting graph structure and rendering it usable by self-attention.

The clearest evidence comes from IMDB-B and IMDB-M, two social-network benchmarks without node attributes. In such settings, a structure-agnostic Transformer has little access to informative graph structure unless the positional encoding itself induces a meaningful geometry over nodes. The substantial advantage of CIPE over LapPE, RWSE, CycleSE, and HKSE on these datasets suggests that communicability provides a graph-wide notion of structural relatedness whose induced similarity geometry is especially well matched to self-attention when positional information is the dominant structural signal.

This advantage extends beyond nearly purely structural benchmarks to PROTEINS, ENZYMES, and the QM7/8/9 molecular regression tasks. 
The strong gains on ENZYMES and QM9 suggest that CIPE remains effective when prediction depends on richer nonlocal or scaffold-level dependencies. 
Together, these results support CIPE as an encoding that captures multi-path graph relatedness and presents it in a form directly usable by self-attention.


\subsection{Positional encoding on Structure-Biased Transformer}

We next consider a stronger setting in which the backbone already incorporates structural inductive bias through message passing. Specifically, we evaluate CIPE on a structure-biased Message-Passing Transformer (MPTr), where a GINE layer first injects local neighborhood information before the Transformer layers. As shown in Table~\ref{tab:mptr_molnet}, CIPE remains consistently beneficial in this setting and emerges as the strongest positional encoding choice for the MPTr backbone across all seven MoleculeNet benchmarks.

\begin{table*}[ht] 
\centering
\caption{Performance on MoleculeNet datasets under the scaffold split. All scores are ROC-AUC ($\uparrow$). Results for MPTr-based variants are reported as mean $\pm$ standard deviation over 5 runs. External baseline results are collected from prior work. Best results are highlighted in bold.}
\label{tab:mptr_molnet}
\resizebox{\linewidth}{!}{
\begin{tabular}{@{}lcccccccc@{}}
\toprule
Model & BACE & BBBP & ClinTox & SIDER & Tox21 & HIV & MUV \\
No. molecules & 1513 & 2039 & 1478 & 1427 & 7831 & 41127 & 93808 \\
Avg. atoms & 65 & 46 & 50.58 & 65 & 36 & 46 & 43 \\
No. tasks & 1 & 1 & 2 & 27 & 12 & 1 & 17 \\
\midrule
SSL pre-training~\cite{hu2020strategies} & 0.799$_{(0.009)}$ & 0.688$_{(0.008)}$ & 0.718$_{(0.041)}$ & 0.610$_{(0.007)}$ & 0.767$_{(0.004)}$ & 0.773$_{(0.010)}$ & 0.758$_{(0.017)}$ \\
GraphCL pre-trained~\cite{you2020graph} 
& 0.754$_{(0.014)}$ & 0.697$_{(0.007)}$ & 0.760$_{(0.027)}$ & 0.605$_{(0.009)}$ & 0.739$_{(0.007)}$ & 0.785$_{(0.012)}$ & 0.698$_{(0.027)}$ \\
InfoGraph pre-trained~\cite{wang2023evaluating} 
& 0.648$_{(0.008)}$ & 0.663$_{(0.006)}$ & 0.663$_{(0.006)}$ & 0.571$_{(0.008)}$ & 0.681$_{(0.006)}$ & 0.702$_{(0.006)}$ & 0.443$_{(0.006)}$ \\
JOAOv2 pre-trained~\cite{wang2023evaluating} 
& 0.674$_{(0.007)}$ & 0.664$_{(0.009)}$ & 0.645$_{(0.009)}$ & 0.591$_{(0.007)}$ & 0.682$_{(0.008)}$ & 0.684$_{(0.005)}$ & 0.474$_{(0.008)}$ \\
GraphMAE~\cite{hou2022graphmae} 
& 0.831$_{(0.009)}$ & 0.720$_{(0.006)}$ & 0.823$_{(0.012)}$ & 0.603$_{(0.011)}$ & 0.755$_{(0.006)}$ & 0.772$_{(0.010)}$ & 0.763$_{(0.024)}$ \\
GraphLoG~\cite{xu2021self} 
& 0.835$_{(0.012)}$ & 0.725$_{(0.008)}$ & 0.767$_{(0.033)}$ & 0.612$_{(0.011)}$ & 0.757$_{(0.005)}$ & 0.778$_{(0.008)}$ & 0.760$_{(0.011)}$ \\
GraphGPS~\cite{rampavsek2022recipe} 
& 0.715$_{(0.060)}$ & 0.562$_{(0.044)}$ & 0.792$_{(0.036)}$ & 0.602$_{(0.011)}$ & 0.714$_{(0.007)}$ & 0.660$_{(0.094)}$ & 0.652$_{(0.016)}$ \\
AttentiveFP~\cite{xiong2019pushing} & 0.784$_{(0.022)}$ & 0.663$_{(0.018)}$ & 0.847$_{(0.003)}$ & 0.606$_{(0.032)}$ & 0.781$_{(0.005)}$ & 0.757$_{(0.014)}$ & \textbf{0.786}$_{(0.015)}$ \\
G2PT~\cite{chen2025graph}
& 0.823$_{(0.016)}$ & 0.710$_{(0.004)}$ & 0.821$_{(0.011)}$ & 0.619$_{(0.002)}$ & 0.750$_{(0.003)}$ & 0.763$_{(0.004)}$ & 0.745$_{(0.003)}$ \\
GPSE$_{\text{augmented}}$~\cite{canturk2024graph} 
& 0.808$_{(0.031)}$ & 0.662$_{(0.009)}$ & 0.788$_{(0.038)}$ & 0.611$_{(0.016)}$ & 0.774$_{(0.008)}$ & 0.772$_{(0.015)}$ & 0.766$_{(0.012)}$ \\
\midrule
MPTr & 0.792$_{(0.013)}$ & 0.709$_{(0.008)}$ & 0.977$_{(0.003)}$ & 0.835$_{(0.002)}$ & 0.776$_{(0.007)}$ & 0.758$_{(0.015)}$ & {0.701}$_{(0.012)}$ \\
MPTr+LapPE & 0.808$_{(0.015)}$ & 0.660$_{(0.012)}$ & 0.976$_{(0.002)}$ & 0.838$_{(0.003)}$ & 0.777$_{(0.006)}$ & 0.760$_{(0.015)}$ & {0.695}$_{(0.021)}$ \\
MPTr+RWSE & 0.795$_{(0.021)}$ & 0.717$_{(0.016)}$ & 0.977$_{(0.002)}$ & 0.839$_{(0.003)}$ & 0.778$_{(0.006)}$ & 0.757$_{(0.011)}$ & {0.699}$_{(0.022)}$ \\
MPTr+CycleSE & 0.801$_{(0.015)}$ & 0.713$_{(0.011)}$ & 0.976$_{(0.001)}$ & 0.842$_{(0.003)}$ & 0.780$_{(0.002)}$ & 0.773$_{(0.013)}$ & {0.696}$_{(0.010)}$ \\
MPTr+HKSE & 0.746$_{(0.078)}$ & 0.658$_{(0.016)}$ & 0.978$_{(0.002)}$ & 0.843$_{(0.004)}$ & 0.789$_{(0.009)}$ & 0.776$_{(0.010)}$ & {0.709}$_{(0.010)}$ \\
\midrule
MPTr+CIPE (ours) & \textbf{0.855}$_{(0.007)}$ & \textbf{0.766}$_{(0.005)}$ & \textbf{0.984}$_{(0.002)}$ & \textbf{0.849}$_{(0.003)}$ & \textbf{0.795}$_{(0.004)}$ & \textbf{0.799}$_{(0.012)}$ & {0.718}$_{(0.018)}$ \\
\bottomrule
\end{tabular}
}
\end{table*}

Overall, MPTr consistently benefits from CIPE even though it already contains structural inductive bias. The gains are largest on BACE, BBBP, and HIV, remain visible on Tox21 and MUV, and are smaller on ClinTox and SIDER. This pattern suggests that CIPE is most beneficial when prediction depends on structural information that is not fully captured by local message passing alone, while smaller gains arise when the baseline is already near saturation or when local structural cues account for much of the predictive signal.

Importantly, CIPE also yields larger gains on MPTr than LapPE, RWSE, CycleSE, or HKSE, whose effects on this structure-biased backbone are typically small and sometimes negative. 
This contrast suggests that CIPE provides structural information complementary to message passing rather than redundant with it. 
By encoding graph-wide multi-path relatedness as an inner-product geometry, CIPE supplies a signal that local neighborhood aggregation does not fully capture and that self-attention can directly exploit.

Beyond the MPTr variants, Table~\ref{tab:mptr_molnet} also compares our method with a broader set of widely used molecular graph models, including GNN-based methods, self-supervised pre-training baselines, and graph Transformer-based models. This comparison shows that the advantage of CIPE is not limited to controlled PE-vs-PE comparisons on the same backbone: MPTr+CIPE remains broadly competitive with these established baselines and achieves stronger performance on multiple datasets. These results further support CIPE as a lightweight yet principled mechanism for injecting complementary global structural information in a form directly usable by self-attention.

\subsection{Sensitivity Analysis}\label{sec:ablation}

We examine the sensitivity of MPTr+CIPE to two key hyperparameters: the positional dimension $d_{\mathrm{pe}}$ and the diffusion time $t$. As shown in Table~\ref{tab:ablation_CIPE_dim_t}, performance varies smoothly across the tested ranges, indicating that CIPE is robust to both the degree of dimensional compression and the structural scale induced by diffusion. 

MPTr+CIPE is generally stable with respect to the positional dimension $d_{\mathrm{pe}}$. As shown in Table~\ref{tab:ablation_CIPE_dim_t}, the setting $d_{\mathrm{pe}}{=}64$ serves as a strong default across MoleculeNet, achieving the best ROC-AUC on six of the seven datasets, while $d_{\mathrm{pe}}{=}32$ causes only modest degradation in most cases. 
BACE is the main exception, where the best result is obtained at $d_{\mathrm{pe}}{=}128$, consistent with its larger average graph size. 
Since dimensionality alignment aims to preserve diffusion-induced inner products after projection, larger graphs may require a larger target dimension to reduce compression-induced distortion.

\begin{table*}[ht]
\centering
\caption{Sensitivity analysis of CIPE on MoleculeNet (ROC-AUC $\uparrow$) under the MPTr backbone. Left block varies the positional encoding dimension $d_{\mathrm{pe}}$, and right block varies the diffusion time $t$. Results are reported as mean $\pm$ std over 5 runs under the same training budget. Best results are highlighted in bold.}
\label{tab:ablation_CIPE_dim_t}
\resizebox{\linewidth}{!}{
\begin{tabular}{@{}lcccccc@{}}
\toprule
& \multicolumn{3}{c}{$d_{\mathrm{pe}}$} & \multicolumn{3}{c}{$t$} \\
\cmidrule(lr){2-4}\cmidrule(lr){5-7}
Dataset & 32 & 64 & 128 & 0.25 & 0.50 & 0.75 \\
\midrule
BACE    & {0.815}$_{(0.017)}$ & {0.822}$_{(0.013)}$ & \textbf{0.855}$_{(0.007)}$ & {0.837}$_{(0.011)}$ & {0.842}$_{(0.007)}$ & \textbf{0.855}$_{(0.007)}$ \\
BBBP    & {0.667}$_{(0.016)}$ & \textbf{0.766}$_{(0.005)}$ & {0.705}$_{(0.012)}$ & {0.671}$_{(0.015)}$ & \textbf{0.766}$_{(0.005)}$ & {0.650}$_{(0.023)}$ \\
ClinTox & {0.977}$_{(0.003)}$ & \textbf{0.984}$_{(0.002)}$ & {0.980}$_{(0.002)}$ & {0.979}$_{(0.003)}$ & {0.977}$_{(0.006)}$ & \textbf{0.984}$_{(0.002)}$ \\
SIDER   & {0.841}$_{(0.004)}$ & \textbf{0.849}$_{(0.003)}$ & {0.837}$_{(0.001)}$ & {0.842}$_{(0.005)}$ & \textbf{0.849}$_{(0.003)}$ & {0.841}$_{(0.004)}$ \\
Tox21   & {0.789}$_{(0.005)}$ & \textbf{0.795}$_{(0.010)}$ & {0.787}$_{(0.006)}$ & {0.790}$_{(0.008)}$ & \textbf{0.795}$_{(0.010)}$ & {0.788}$_{(0.004)}$ \\
HIV     & {0.787}$_{(0.011)}$ & \textbf{0.799}$_{(0.012)}$ & {0.789}$_{(0.016)}$ & {0.789}$_{(0.008)}$ & \textbf{0.799}$_{(0.012)}$ & {0.790}$_{(0.013)}$ \\
MUV     & {0.707}$_{(0.010)}$ & \textbf{0.718}$_{(0.018)}$ & {0.712}$_{(0.012)}$ & {0.707}$_{(0.034)}$ & {0.705}$_{(0.025)}$ & \textbf{0.718}$_{(0.018)}$ \\
\bottomrule
\end{tabular}
}
\end{table*}

MPTr+CIPE is also robust to the diffusion time $t$, with $t{=}0.5$ providing a reliable default across datasets. The diffusion time controls the structural scale encoded by CIPE: smaller values emphasize shorter-range connectivity, whereas larger values aggregate evidence over longer and more indirect paths. Table~\ref{tab:ablation_CIPE_dim_t} shows that $t{=}0.5$ achieves the best performance on BBBP, SIDER, Tox21, and HIV, while remaining close to optimal on the remaining benchmarks. Meanwhile, BACE, ClinTox, and MUV attain their best scores at $t{=}0.75$, suggesting that larger or structurally more involved graphs may benefit from slightly deeper diffusion. Overall, the variation across $t\in\{0.25,0.5,0.75\}$ remains modest, indicating that CIPE is not overly sensitive to the precise diffusion scale and can be tuned effectively with a coarse grid.

Taken together, these results indicate that CIPE is practically robust with respect to both embedding dimension and diffusion scale. 
A moderate target dimension ($d_{\mathrm{pe}}{=}64$) and a middle-range diffusion time ($t{=}0.5$) are sufficient in most cases, suggesting that CIPE can be tuned with coarse hyperparameter choices rather than finely optimized settings. 
Additional sensitivity analyses with respect to the learning rate, batch size, and Transformer hidden dimension are reported in the Supplementary Information, Tables S3--S5.

\section{Conclusion}

We presented Communicability-Inspired Positional Encoding (CIPE), a graph positional encoding framework that goes beyond encoding structural information as auxiliary node descriptors. 
CIPE is designed to align graph structural representations with the inner-product similarity mechanism of self-attention. 
By deriving node encodings from diffusion-based communicability, CIPE induces a positional geometry in which pairwise inner products directly reflect global, multi-path structural relatedness between vertices. 
This construction provides an attention-compatible way to inject graph-wide connectivity into Transformer models.

To support practical use across graphs of different sizes, we further introduced a dimensionality alignment procedure for mapping graph-size-dependent CIPE representations to a prescribed embedding dimension. 
The alignment is formulated to preserve the communicability-induced inner-product geometry as faithfully as possible, yielding a fixed-dimensional and numerically stable positional encoding for graph Transformers. 
Experiments on 14 benchmarks covering molecular property prediction, quantum-chemical regression, and social-network and bio/protein graph classification demonstrate the effectiveness of CIPE. 
CIPE achieves substantial improvements on structure-agnostic Transformers, where positional encoding serves as the primary source of graph structure, and also provides consistent gains on structure-biased graph Transformers, indicating its complementarity to local message passing.

Overall, this work establishes an operator-driven paradigm for graph positional encoding in Transformers. 
It highlights that effective graph PEs should not only capture informative structural signals, but also organize them into geometries compatible with attention, with attention-compatible inner-product structure being a central design criterion. 
This perspective may inform future positional encoding designs based on other domain-relevant operators and broader classes of structured data.

\section*{Acknowledgments}
This work was supported in part by the Singapore Ministry of Education Academic Research Fund Tier 1 grant RG16/23, Tier 2 grants MOE-T2EP20125-0004.


\bibliographystyle{IEEEtran}
\bibliography{references}

\begin{thebibliography}{10}
\providecommand{\url}[1]{#1}
\csname url@samestyle\endcsname
\providecommand{\newblock}{\relax}
\providecommand{\bibinfo}[2]{#2}
\providecommand{\BIBentrySTDinterwordspacing}{\spaceskip=0pt\relax}
\providecommand{\BIBentryALTinterwordstretchfactor}{4}
\providecommand{\BIBentryALTinterwordspacing}{\spaceskip=\fontdimen2\font plus
\BIBentryALTinterwordstretchfactor\fontdimen3\font minus \fontdimen4\font\relax}
\providecommand{\BIBforeignlanguage}[2]{{%
\expandafter\ifx\csname l@#1\endcsname\relax
\typeout{** WARNING: IEEEtran.bst: No hyphenation pattern has been}%
\typeout{** loaded for the language `#1'. Using the pattern for}%
\typeout{** the default language instead.}%
\else
\language=\csname l@#1\endcsname
\fi
#2}}
\providecommand{\BIBdecl}{\relax}
\BIBdecl

\bibitem{vaswani2017attention}
A.~Vaswani, N.~Shazeer, N.~Parmar, J.~Uszkoreit, L.~Jones, A.~N. Gomez, {\L}.~Kaiser, and I.~Polosukhin, ``Attention is all you need,'' \emph{Advances in neural information processing systems}, vol.~30, 2017.

\bibitem{tang2018self}
G.~Tang, M.~M{\"u}ller, A.~R. Gonzales, and R.~Sennrich, ``Why self-attention? a targeted evaluation of neural machine translation architectures,'' in \emph{Proceedings of the 2018 conference on empirical methods in natural language processing}, 2018, pp. 4263--4272.

\bibitem{devlin2019bert}
J.~Devlin, M.-W. Chang, K.~Lee, and K.~Toutanova, ``Bert: Pre-training of deep bidirectional transformers for language understanding,'' in \emph{Proceedings of the 2019 conference of the North American chapter of the association for computational linguistics: human language technologies, volume 1 (long and short papers)}, 2019, pp. 4171--4186.

\bibitem{gillioz2020overview}
A.~Gillioz, J.~Casas, E.~Mugellini, and O.~Abou~Khaled, ``Overview of the transformer-based models for nlp tasks,'' in \emph{2020 15th Conference on computer science and information systems (FedCSIS)}.\hskip 1em plus 0.5em minus 0.4em\relax IEEE, 2020, pp. 179--183.

\bibitem{dosovitskiy2020image}
A.~Dosovitskiy, L.~Beyer, A.~Kolesnikov, D.~Weissenborn, X.~Zhai, T.~Unterthiner, M.~Dehghani, M.~Minderer, G.~Heigold, S.~Gelly \emph{et~al.}, ``An image is worth 16x16 words: Transformers for image recognition at scale,'' \emph{arXiv preprint arXiv:2010.11929}, 2020.

\bibitem{khan2022transformers}
S.~Khan, M.~Naseer, M.~Hayat, S.~W. Zamir, F.~S. Khan, and M.~Shah, ``Transformers in vision: A survey,'' \emph{ACM computing surveys (CSUR)}, vol.~54, no. 10s, pp. 1--41, 2022.

\bibitem{shaw2018self}
P.~Shaw, J.~Uszkoreit, and A.~Vaswani, ``Self-attention with relative position representations,'' \emph{arXiv preprint arXiv:1803.02155}, 2018.

\bibitem{raffel2020exploring}
C.~Raffel, N.~Shazeer, A.~Roberts, K.~Lee, S.~Narang, M.~Matena, Y.~Zhou, W.~Li, and P.~J. Liu, ``Exploring the limits of transfer learning with a unified text-to-text transformer,'' \emph{Journal of machine learning research}, vol.~21, no. 140, pp. 1--67, 2020.

\bibitem{press2022train}
\BIBentryALTinterwordspacing
O.~Press, N.~Smith, and M.~Lewis, ``Train short, test long: Attention with linear biases enables input length extrapolation,'' in \emph{International Conference on Learning Representations}, 2022. [Online]. Available: \url{https://openreview.net/forum?id=R8sQPpGCv0}
\BIBentrySTDinterwordspacing

\bibitem{su2024roformer}
J.~Su, M.~Ahmed, Y.~Lu, S.~Pan, W.~Bo, and Y.~Liu, ``Roformer: Enhanced transformer with rotary position embedding,'' \emph{Neurocomputing}, vol. 568, p. 127063, 2024.

\bibitem{dufter2022position}
P.~Dufter, M.~Schmitt, and H.~Sch{\"u}tze, ``Position information in transformers: An overview,'' \emph{Computational Linguistics}, vol.~48, no.~3, pp. 733--763, 2022.

\bibitem{bronstein2021geometric}
M.~M. Bronstein, J.~Bruna, T.~Cohen, and P.~Veli{\v{c}}kovi{\'c}, ``Geometric deep learning: Grids, groups, graphs, geodesics, and gauges,'' \emph{arXiv preprint arXiv:2104.13478}, 2021.

\bibitem{shervashidze2011weisfeiler}
N.~Shervashidze, P.~Schweitzer, E.~J. Van~Leeuwen, K.~Mehlhorn, and K.~M. Borgwardt, ``Weisfeiler-lehman graph kernels.'' \emph{Journal of Machine Learning Research}, vol.~12, no.~9, 2011.

\bibitem{henderson2011s}
K.~Henderson, B.~Gallagher, L.~Li, L.~Akoglu, T.~Eliassi-Rad, H.~Tong, and C.~Faloutsos, ``It's who you know: graph mining using recursive structural features,'' in \emph{Proceedings of the 17th ACM SIGKDD international conference on Knowledge discovery and data mining}, 2011, pp. 663--671.

\bibitem{henderson2012rolx}
K.~Henderson, B.~Gallagher, T.~Eliassi-Rad, H.~Tong, S.~Basu, L.~Akoglu, D.~Koutra, C.~Faloutsos, and L.~Li, ``Rolx: structural role extraction \& mining in large graphs,'' in \emph{Proceedings of the 18th ACM SIGKDD international conference on Knowledge discovery and data mining}, 2012, pp. 1231--1239.

\bibitem{dwivedi2023benchmarking}
V.~P. Dwivedi, C.~K. Joshi, A.~T. Luu, T.~Laurent, Y.~Bengio, and X.~Bresson, ``Benchmarking graph neural networks,'' \emph{Journal of Machine Learning Research}, vol.~24, no.~43, pp. 1--48, 2023.

\bibitem{canturk2023graph}
S.~Cant{\"u}rk, R.~Liu, O.~Lapointe-Gagn{\'e}, V.~L{\'e}tourneau, G.~Wolf, D.~Beaini, and L.~Ramp{\'a}{\v{s}}ek, ``Graph positional and structural encoder,'' \emph{arXiv preprint arXiv:2307.07107}, 2023.

\bibitem{yan2024cycle}
Z.~Yan, T.~Ma, L.~Gao, Z.~Tang, C.~Chen, and Y.~Wang, ``Cycle invariant positional encoding for graph representation learning,'' in \emph{Learning on Graphs Conference}.\hskip 1em plus 0.5em minus 0.4em\relax PMLR, 2024, pp. 4--1.

\bibitem{dwivedi2020generalization}
V.~P. Dwivedi and X.~Bresson, ``A generalization of transformer networks to graphs,'' \emph{arXiv preprint arXiv:2012.09699}, 2020.

\bibitem{huang2023stability}
Y.~Huang, W.~Lu, J.~Robinson, Y.~Yang, M.~Zhang, S.~Jegelka, and P.~Li, ``On the stability of expressive positional encodings for graphs,'' \emph{arXiv preprint arXiv:2310.02579}, 2023.

\bibitem{kreuzer2021rethinking}
D.~Kreuzer, D.~Beaini, W.~Hamilton, V.~L{\'e}tourneau, and P.~Tossou, ``Rethinking graph transformers with spectral attention,'' \emph{Advances in Neural Information Processing Systems}, vol.~34, pp. 21\,618--21\,629, 2021.

\bibitem{mialon2021graphit}
G.~Mialon, D.~Chen, M.~Selosse, and J.~Mairal, ``Graphit: Encoding graph structure in transformers,'' \emph{arXiv preprint arXiv:2106.05667}, 2021.

\bibitem{rampavsek2022recipe}
L.~Ramp{\'a}{\v{s}}ek, M.~Galkin, V.~P. Dwivedi, A.~T. Luu, G.~Wolf, and D.~Beaini, ``Recipe for a general, powerful, scalable graph transformer,'' \emph{Advances in Neural Information Processing Systems}, vol.~35, pp. 14\,501--14\,515, 2022.

\bibitem{min2022transformer}
E.~Min, R.~Chen, Y.~Bian, T.~Xu, K.~Zhao, W.~Huang, P.~Zhao, J.~Huang, S.~Ananiadou, and Y.~Rong, ``Transformer for graphs: An overview from architecture perspective,'' \emph{arXiv preprint arXiv:2202.08455}, 2022.

\bibitem{corso2024graph}
G.~Corso, H.~Stark, S.~Jegelka, T.~Jaakkola, and R.~Barzilay, ``Graph neural networks,'' \emph{Nature Reviews Methods Primers}, vol.~4, no.~1, p.~17, 2024.

\bibitem{leicht2006vertex}
E.~A. Leicht, P.~Holme, and M.~E. Newman, ``Vertex similarity in networks,'' \emph{Physical Review E—Statistical, Nonlinear, and Soft Matter Physics}, vol.~73, no.~2, p. 026120, 2006.

\bibitem{pereda2019visualization}
M.~Pereda and E.~Estrada, ``Visualization and machine learning analysis of complex networks in hyperspherical space,'' \emph{Pattern Recognition}, vol.~86, pp. 320--331, 2019.

\bibitem{estrada2024communicability}
E.~Estrada, ``Communicability cosine distance: similarity and symmetry in graphs/networks,'' \emph{Computational and Applied Mathematics}, vol.~43, no.~1, p.~49, 2024.

\bibitem{estrada2008communicability}
E.~Estrada and N.~Hatano, ``Communicability in complex networks,'' \emph{Physical Review E—Statistical, Nonlinear, and Soft Matter Physics}, vol.~77, no.~3, p. 036111, 2008.

\bibitem{estrada2011structure}
E.~Estrada, \emph{The Structure of Complex Networks: Theory and Applications}.\hskip 1em plus 0.5em minus 0.4em\relax New York: OUP Oxford, 2011.

\bibitem{grindrod2011communicability}
P.~Grindrod, M.~C. Parsons, D.~J. Higham, and E.~Estrada, ``Communicability across evolving networks,'' \emph{Physical Review E}, vol.~83, no.~4, p. 046120, 2011.

\bibitem{estrada2014hyperspherical}
E.~Estrada, M.~Sanchez-Lirola, and J.~A. De~La~Pe{\~n}a, ``Hyperspherical embedding of graphs and networks in communicability spaces,'' \emph{Discrete Applied Mathematics}, vol. 176, pp. 53--77, 2014.

\bibitem{estrada2012physics}
E.~Estrada, N.~Hatano, and M.~Benzi, ``The physics of communicability in complex networks,'' \emph{Physics reports}, vol. 514, no.~3, pp. 89--119, 2012.

\bibitem{estrada2016communicability}
E.~Estrada and N.~Hatano, ``Communicability angle and the spatial efficiency of networks,'' \emph{SIAM Review}, vol.~58, no.~4, pp. 692--715, 2016.

\bibitem{wu2018moleculenet}
Z.~Wu, B.~Ramsundar, E.~N. Feinberg, J.~Gomes, C.~Geniesse, A.~S. Pappu, K.~Leswing, and V.~Pande, ``Moleculenet: a benchmark for molecular machine learning,'' \emph{Chemical science}, vol.~9, no.~2, pp. 513--530, 2018.

\bibitem{morris2020tudataset}
C.~Morris, N.~M. Kriege, F.~Bause, K.~Kersting, P.~Mutzel, and M.~Neumann, ``Tudataset: A collection of benchmark datasets for learning with graphs,'' in \emph{ICML 2020 Workshop on Graph Representation Learning and Beyond (GRL+ 2020)}, 2020.

\bibitem{estrada2009communicabilitygraph}
E.~Estrada and N.~Hatano, ``Communicability graph and community structures in complex networks,'' \emph{Applied Mathematics and Computation}, vol. 214, no.~2, pp. 500--511, 2009.

\bibitem{estrada2009communicability}
E.~Estrada, D.~J. Higham, and N.~Hatano, ``Communicability betweenness in complex networks,'' \emph{Physica A: Statistical Mechanics and its Applications}, vol. 388, no.~5, pp. 764--774, 2009.

\bibitem{estrada2012complex}
E.~Estrada, ``Complex networks in the euclidean space of communicability distances,'' \emph{Physical Review E—Statistical, Nonlinear, and Soft Matter Physics}, vol.~85, no.~6, p. 066122, 2012.

\bibitem{hammond2011wavelets}
D.~K. Hammond, P.~Vandergheynst, and R.~Gribonval, ``Wavelets on graphs via spectral graph theory,'' \emph{Applied and computational harmonic analysis}, vol.~30, no.~2, pp. 129--150, 2011.

\bibitem{al2011computing}
A.~H. Al-Mohy and N.~J. Higham, ``Computing the action of the matrix exponential, with an application to exponential integrators,'' \emph{SIAM journal on scientific computing}, vol.~33, no.~2, pp. 488--511, 2011.

\bibitem{hu2020strategies}
W.~Hu, B.~Liu, J.~Gomes, M.~Zitnik, P.~Liang, V.~Pande, and J.~Leskovec, ``Strategies for pre-training graph neural networks,'' in \emph{International Conference on Learning Representations (ICLR)}, 2020.

\bibitem{you2020graph}
Y.~You, T.~Chen, Y.~Sui, T.~Chen, Z.~Wang, and Y.~Shen, ``Graph contrastive learning with augmentations,'' \emph{Advances in neural information processing systems}, vol.~33, pp. 5812--5823, 2020.

\bibitem{wang2023evaluating}
H.~Wang, J.~Kaddour, S.~Liu, J.~Tang, J.~Lasenby, and Q.~Liu, ``Evaluating self-supervised learning for molecular graph embeddings,'' \emph{Advances in Neural Information Processing Systems}, vol.~36, pp. 68\,028--68\,060, 2023.

\bibitem{hou2022graphmae}
Z.~Hou, X.~Liu, Y.~Cen, Y.~Dong, H.~Yang, C.~Wang, and J.~Tang, ``Graphmae: Self-supervised masked graph autoencoders,'' in \emph{Proceedings of the 28th ACM SIGKDD conference on knowledge discovery and data mining}, 2022, pp. 594--604.

\bibitem{xu2021self}
M.~Xu, H.~Wang, B.~Ni, H.~Guo, and J.~Tang, ``Self-supervised graph-level representation learning with local and global structure,'' in \emph{International conference on machine learning}.\hskip 1em plus 0.5em minus 0.4em\relax PMLR, 2021, pp. 11\,548--11\,558.

\bibitem{xiong2019pushing}
Z.~Xiong, D.~Wang, X.~Liu, F.~Zhong, X.~Wan, X.~Li, Z.~Li, X.~Luo, K.~Chen, H.~Jiang \emph{et~al.}, ``Pushing the boundaries of molecular representation for drug discovery with the graph attention mechanism,'' \emph{Journal of Medicinal Chemistry}, vol.~63, no.~16, pp. 8749--8760, 2019.

\bibitem{chen2025graph}
X.~Chen, Y.~Wang, J.~He, Y.~Du, S.~Hassoun, X.~Xu, and L.~Liu, ``Graph generative pre-trained transformer,'' in \emph{International Conference on Machine Learning}.\hskip 1em plus 0.5em minus 0.4em\relax PMLR, 2025, pp. 9176--9197.

\bibitem{canturk2024graph}
S.~Cant{\"u}rk, R.~Liu, O.~Lapointe-Gagn{\'e}, V.~L{\'e}tourneau, G.~Wolf, D.~Beaini, and L.~Ramp{\'a}{\v{s}}ek, ``Graph positional and structural encoder,'' in \emph{Proceedings of the 41st International Conference on Machine Learning}, 2024, pp. 5533--5566.

\end{thebibliography}


 




\vfill

\end{document}